\documentclass[sigconf]{acmart}
\AtBeginDocument{%
  \providecommand\BibTeX{{%
    \normalfont B\kern-0.5em{\scshape i\kern-0.25em b}\kern-0.8em\TeX}}}

\setcopyright{acmcopyright}
\copyrightyear{}
\acmYear{2020}
\acmDOI{}

\acmConference[KG-Bias '20]{KG-Bias '20: Bias in Automatic Knowledge Graph Construction}{June 24, 2020}{Virtual}
\acmBooktitle{Proceedings of the Bias in Automatic Knowledge Graph Construction - A Workshop at AKBC 2020,
  June 24, 2020, Virtual}
\acmPrice{}
\acmISBN{}

\usepackage{graphicx}
\usepackage{booktabs}
\usepackage{enumitem}

\usepackage{subcaption}
\usepackage{color, colortbl}

\definecolor{Gray}{gray}{0.9}
\newcommand{\ControlName}{\textbf{Syedtiastephen}}

\setcopyright{none}
\begin{document}

\title{Assessing Demographic Bias in Named Entity Recognition}

\author{Shubhanshu Mishra}
\authornote{All authors contributed equally to this research.}
\email{smishra@twitter.com}
\orcid{1234-5678-9012}
\affiliation{%
  \institution{Twitter, Inc.}
}

\author{Sijun He}
\authornotemark[1]
\email{she@twitter.com}
\affiliation{%
  \institution{Twitter, Inc.}
}

\author{Luca Belli}
\authornotemark[1]
\email{lbelli@twitter.com}

\affiliation{%
  \institution{Twitter, Inc.}
}

% \affiliation{%

% \affiliation{%

% \affiliation{%

% \affiliation{%

% \affiliation{%



\renewcommand{\shortauthors}{Mishra, He, and Belli}

\begin{abstract}
  Named Entity Recognition (NER) is often the first step towards automated Knowledge Base (KB) generation from raw text. In this work, we assess the bias in various Named Entity Recognition (NER) systems for English across different demographic groups with synthetically generated corpora. Our analysis reveals that models perform better at identifying names from specific demographic groups across two datasets. We also identify that debiased embeddings do not help in resolving this issue. Finally, we observe that character-based contextualized word representation models such as ELMo results in the least bias across demographics. Our work can shed light on potential biases in automated KB generation due to systematic exclusion of named entities belonging to certain demographics.
\end{abstract}





\maketitle

\section{Introduction}

In recent times, there has been growing interest around bias in algorithmic decision making and machine learning systems, especially on how automated decisions are affecting different segments of the population and can amplify or exacerbate existing biases in society \cite{safiya}. While many of the NLP ethics research papers focus on understanding and mitigating the bias present in embeddings \cite{Bolukbasi2016, gonen-goldberg-2019-lipstick-pig}, bias in Named Entity Recognition (NER) \cite{tjong-kim-sang-de-meulder-2003-introduction,mishra-diesner-2016-semi,Mishra2019} is not scrutinized in the same way. NER is widely-used as the first step of a variety of NLP applications, ranging from large-scale search systems \cite{Pujara2018wsdmKGtutorial} to automated knowledge graphs (KG) and knowledge base (KB) generation \cite{Guo2019SIGIRtutorialSearch}. Bias in the first step of a pipeline could propagate throughout the entire system, leading to allocation and representation harm \cite{Barocas2017Crawford}. 

% Allocation harm is due the lack of opportunity (e.g. online exposure) that is offered to others. This lack of fair representation will make part of the population invisible for future data collection jobs (e.g. to retrain models) thus reinforcing the problem. 

While most prior work focused on bias in embeddings, previous work has not given much attention to bias in NER systems. Understanding bias in NER systems is essential as these systems are used for several downstream NLP applications. To fill this gap, we analyze the bias in commonly used NER systems.

In this work, we analyze widely-used NER models to identify demographic bias when performing NER. We seek to answer the following question: \emph{Other things held constant, are names commonly associated with certain demographic categories like genders or ethnicities more likely to be recognized?}

Our contributions in this paper are the following:
\begin{enumerate}[noitemsep]
    \item Propose a novel framework\footnote{Details will be available at: \href{https://github.com/napsternxg/NER\_bias}{https://github.com/napsternxg/NER\_bias}} to analyze bias in NER systems, including a methodology for creating a synthetic dataset using a small seed list of names.
    \item Show that there exists systematic bias of existing NER methods in failing to identify named entities from certain demographics.
\end{enumerate}

\section{Experimental Setup}

Our general experimental setup is based on using synthetically generated data to assess the bias in common NER models, which includes popular NER model architectures trained on standard datasets and off-the-shelf models from commonly-used NLP libraries. As discussed in Section \ref{data-gen}, we create the dataset with controlled context so that the effect of the names are properly marginalized and measured. We perform inference with various models on the dataset to extract person named entities and measure the respective accuracy and confidence of the correctly extracted names. Since capitalization is considered as an important feature for NER, we repeat the experiment with and without the capitalization of the name.
% \begin{enumerate}[noitemsep]
%     \item Was the name recognized?
% \item (if available) What was the confidence of the prediction (0 if not recognized)?
% \item What were the original metadata of the name?
% \end{enumerate}

\subsection{Data Generation and Pre-processing} \label{data-gen}

\begin{table}[!htbp]
    \centering
    \begin{tabular}{p{0.25\linewidth}p{0.68\linewidth}}
\toprule
\textbf{category} & \textbf{Names}\\
\midrule
\textbf{Black Female (BF)      } &  Aaliyah, Ebony, Jasmine, Lakisha, Latisha, Latoya, Malika, Nichelle, Nishelle, Shanice, Shaniqua, Shereen, Tanisha, Tia, Yolanda, Yvette \\
\textbf{Black Male (BM)      } &  Alonzo, Alphonse, Darnell, Deion, Jamel, Jerome, Lamar, Lamont, Leroy, Lionel, Malik, Terrence, Theo, Torrance, Tyree \\
\midrule
\textbf{Hispanic Female (HF)      } &  Ana, Camila, Elena, Isabella, Juana, Luciana, Luisa, Maria, Mariana, Martina, Sofia, Valentina, Valeria, Victoria, Ximena \\
\textbf{Hispanic Male (HM)      } &  Alejandro, Daniel, Diego, Jorge, Jose, Juan, Luis, Mateo, Matias, Miguel, Nicolas, Samuel, Santiago, Sebastian, Tomas \\
\midrule
\textbf{Muslim Female (MF)      } &  Alya, Ayesha, Fatima, Jana, Lian, Malak, Mariam, Maryam, Nour, Salma, Sana, Shaista, Zahra, Zara, Zoya \\
\textbf{Muslim Male (MM)      } &  Abdullah, Ahmad, Ahmed, Ali, Ayaan, Hamza, Mohammed, Omar, Rayyan, Rishaan, Samar, Syed, Yasin, Youssef, Zikri \\
\midrule
\textbf{White Female (WF)      } &  Amanda, Betsy, Colleen, Courtney, Ellen, Emily, Heather, Katie, Kristin, Lauren, Megan, Melanie, Nancy, Rachel, Stephanie \\
\textbf{White Male (WM)      } &  Adam, Alan, Andrew, Brad, Frank, Greg, Harry, Jack, Josh, Justin, Matthew, Paul, Roger, Ryan, Stephen \\
\midrule
\textbf{OOV Name} &  \ControlName \\
\bottomrule
\end{tabular}
    \caption{Name lists from different demographics. }
    \label{tab:name_categories}
\end{table}

\begin{table*}[!htbp]
    \centering
    \begin{tabular}{l|p{0.85\textwidth}}
        \toprule
        \multicolumn{2}{c}{\textbf{WINOGENDER}}\\
        \midrule
        Original & \textbf{\$OCCUPATION} told \textbf{\$PARTICIPANT} that \textbf{\$NOM\_PRONOUN} could pay with cash.\\
        Sample 1 & \textbf{Alya} told \textbf{Jasmine} that \textbf{Andrew} could pay with cash.\\
        Sample 2 & \textbf{Alya} told \textbf{Theo} that \textbf{Ryan} could pay with cash.\\
        \midrule
        \multicolumn{2}{c}{\textbf{IN-SITU (CoNLL 03 Test)}}\\
        \midrule
        Original & \textbf{Charlton} managed Ireland for 93 matches , during which time they lost only 17 times in almost 10 years until he resigned in December 1995 .\\
        Sample 1 & \textbf{Syed} managed Ireland for 93 matches , during which time they lost only 17 times in almost 10 years until he resigned in December 1995 .\\
        \bottomrule
    \end{tabular}
    \caption{Examples of synthetic dataset generated from Winogender Schema and CoNLL 03 test data.}
    \label{tab:winogender_examples}
\end{table*}

In order to assess the bias in NER across different demographic groups, we need a corpus of sentences in which the named entity is equally likely to be from either demographic category. We overcome this issue by using sentence templates with placeholders to be filled with different names. In this work we only focus on \textbf{unigram person named entities}. Below we outline our approach for generating named entity corpora from two types of sentence templates. Using the same sentence with different names allows us to remove the confounding effect introduced by the sentence structure.

\textbf{Names.} Our name collection consists of 123 names across 8 different demographic groups, which are a combination of race\footnote{\url{https://www.census.gov/topics/population/race/about.html}} (or ethnicity) and gender. The categories span racial (or ethnic) categories, namely, Black, White, Hispanic, and Muslim \footnote{We include Muslim and Hispanic along with other racial categories to better organize our results. We are aware that they are not racial categories.}. For each race we include two gender categories, namely, male and female. Each demographic category, is represented in our name collection with 15 salient names (and one with 16 names). A detailed list of names and their demographic categories is provided in Table \ref{tab:name_categories}.

Our name collection is constructed from two different sources. The first source of names comes from popular male and female first names among White and Black communities and was used to study the effect of gender bias in resume reviews in the work by \citet{Bertrand2004Gender}. This name dataset was constructed based on the most salient names for each demographic groups among the baby births registered in Massachusetts between 1974 and 1979\footnote{While we are aware that name distributions might have changed slightly in recent years, we think it's a reasonable list for this project}. The second source contains names in all eight demographic categories and is taken from the ConceptNet project\footnote{\url{https://github.com/commonsense/conceptnet5/blob/master/conceptnet5/vectors/evaluation/bias.py}} \cite{speer2017conceptnet}. This collection of names was used to debias the ConceptNet embeddings \cite{sweeney-najafian-2019-transparent}.  %\footnote{\url{http://blog.conceptnet.io/posts/2017/conceptnet-numberbatch-17-04-better-less-stereotyped-word-vectors/}}.
%We utilize the debiased ConceptNet embeddings to later evaluate our NER model to asses if this debiasing helps in reducing NER model bias.
We introduce a baseline name category to measure the context-only performance of the NER models with uninformative embedding. As described later, we also trained a few models in-house, for those models we directly use the OOV token. For pre-trained models, we use \ControlName, which is unlikely to be found in the vocabulary but has the word shape features of a name. Hispanic names were deaccented (i.e. \textit{José} becomes \textit{Jose}) because including the accented names resulted in a higher OOV rate for Hispanic names. 

We are aware that our work is limited by the availability of names from various demographics and we acknowledge that individuals will not-necessarily identity themselves with the demographics attached to their first name, as done in this work. Furthermore, we do not endorse using this name list for inferring any demographic attributes for an individual because the demographic attributes are personal identifiers and this method is error prone when done at an individual level. For the sake of brevity unless explicitly specified, we refer to names in our list by the community they are most likely to be found as specified in table \ref{tab:name_categories}. This means that when we refer to \textbf{White Female Names} we mean names categorized as \textbf{White Female} in table \ref{tab:name_categories}.

Among our name collections, the names \textbf{Nishelle (BF)}, \textbf{Rishaan (MM)}, and \textbf{Zikri (MM)} are not found in  the Stanford GloVe \cite{pennington-etal-2014-glove} embeddings' vocabulary. Furthermore, the names \textbf{Ayaan (MM)}, \textbf{Lakisha (BF)}, \textbf{Latisha (BF)}, \textbf{Nichelle (BF)}, \textbf{Nishelle (BF)}, \textbf{Rishaan (MM)}, and \textbf{Shereen (BF)} are not found in the ConceptNet embedding vocabulary. 

\textbf{Winogender.} Now we describe how to on generate synthetic sentences using sentence templates. We propose to generate synthetic  sentences using the sentences provided by the Winogender Schemas \cite{rudinger-etal-2018-gender} project. The original goal of Winogender Schemas is to find gender bias in automated co-reference solutions. We modify their templates to make them more appropriate for generating synthetic templates using named entities. Our modification included removing the word \textit{the} before the placeholder in the templates and removing templates which have less than 3 placeholders. Examples of the cleaned up template and samples generated by us is shown in Table \ref{tab:winogender_examples}.
We generated samples by replacing instance of  \textit{\$OCCUPATION},  \textit{\$PARTICIPANT} and \textit{\$NOM\_PRONOUN} in the templates with the names in our list, thus stretching their original intent. This gives us syntactically and semantically correct sentences. We utilize all triples of names, for each sentence template resulting in a corpus of $3! * {123 \choose 3} = 217$ million unique sentences. %We are aware of the limitations of the sentence format, i.e. first named entities always appear at the start of the sentence. Hence, in our results we also conduct an analysis focusing only on the names present in the last placeholder i.e. \textit{\$NOM\_PRONOUN}, for which the models are likely to observe less noise and give more accurate predictions.

\textbf{In-Situ.} To investigate the performance of the models on names in real world (or in-situ) data, we synthesize a more realistic dataset by performing name replacement with the CoNLL 2003 NER test data \cite{tjong-kim-sang-de-meulder-2003-introduction}.  Sentences with more than 5 tokens (to ensure proper context) and contain exactly one unigram person entity (see limitations part of Section \ref{discussion}) are selected in this data. As a result, the sentence can have other n-gram entities of all types. This results in a dataset of 289 sentences. We again create synthetic sentences by replacing the unigram PERSON entity with the names described above. %We call this dataset \textbf{In-situ dataset}. 

Finally, we replicate our evaluations on lower-cased data (both Winogender and In-Situ) to investigate how the models perform when the sentences (including the names) are lower-cased; this removes the dominance of word shape features and checks purely for syntactic feature usage. This setting also resembles social media text, where capitalization rules are not very often followed \cite{mishra-diesner-2016-semi,Mishra2019}.

\subsection{Models}
\label{models}
We assessed the bias on the following widely-used NER model architectures as well as off-the-shelf libraries:

\begin{enumerate}[noitemsep]
    \item \textbf{BiLSTM CRF} \cite{bilstmcrf,lample-etal-2016-neural} is one of the most commonly-used deep learning architectures for NER. The model uses pre-trained word embeddings as input representations, bidirectional LSTM to compose context-dependent representations of the text from both directions, and Conditional Random Field (CRF) \cite{crf} to decode output into a sequence of tags. Since we are interested in both the correctness as well as the confidence of extracted named entities, we also compute the entity-level confidence via the \textit{Constrained Forward-Backward algorithm} \cite{culotta-mccallum-2004-confidence}. Different versions of this model were trained on CoNLL 03 NER benchmark dataset \cite{tjong-kim-sang-de-meulder-2003-introduction} by utilizing varying embedding methods:
    
    \begin{enumerate}[noitemsep]
        \item \textbf{GloVe} uses GloVe 840B word vectors pre-trained on Common Crawl \cite{pennington-etal-2014-glove}. 
        \item \textbf{CNET} uses ConceptNet english embeddings (version 1908) \cite{speer2017conceptnet}, which have already been debiased for gender and ethnicity \footnote{\href{https://blog.conceptnet.io/posts/2017/conceptnet-numberbatch-17-04-better-less-stereotyped-word-vectors/}{https://blog.conceptnet.io/posts/2017/conceptnet-numberbatch-17-04-better-less-stereotyped-word-vectors/}}. 
        \item \textbf{ELMo} uses ELMo embeddings \cite{peters-etal-2017-semi}, which provides contextualized representations from a deep bidirectional language model where the words are encoded using embeddings of their characters. This approach allows us to overcome the OOV issue. 
    \end{enumerate}
    \item \textbf{spaCy} is a widely-used open-source library for NLP that features pre-trained NER models. We performed analysis on \textbf{spacy\_sm} and \textbf{spacy\_lg} English NER models from spaCy version 2.1.0\footnote{\url{https://spacy.io/}}. 
    spaCy models are trained on OntoNotes 5\footnote{\url{https://spacy.io/models/en}} data.

    \item \textbf{Stanford CoreNLP} \footnote{\url{https://stanfordnlp.github.io/CoreNLP/history.html}} (\textbf{corenlp}) \cite{manning-etal-2014-stanford} is one of the most popular NLP library and we use the \textit{2018-10-05} version. CoreNLP NER was trained (by its authors) on data from CoNLL03 and ACE 2002\footnote{\url{https://nlp.stanford.edu/software/CRF-NER.html\#Models}}. 
\end{enumerate}

%Current author: Sijun%
\textbf{Note on excluding BERT} While the approach of fine-tuning large pre-trained transformer language models such as BERT \cite{devlin-etal-2019-bert} has established state-of-the-art performance on NER, the implementations used subword tokenization such as WordPiece \cite{wordpiece} or Byte-Pair-Coding \cite{wordpiece} which require pre-tokenization followed by word pieces for NER tasks where the prediction has to be made on the word level.
Although the BERT paper has addressed this issue by using the embedding of the first subword token for each word, this breaks the unigram entity assumption we have used in our analysis. Furthermore, number of BERT tokens may vary for names adding another degree of freedom to control. 
Furthermore, our inclusion of ELMo can be considered as a fair comparison for utilizing contextual word embeddings compared to other models which uses fixed word embeddings.

\subsection{Evaluation Criteria}
The goal of this work is to assess if NER models vary in their accuracy of identifying first names from various demographics as an instance of named entity with label $l=PERSON$. Assuming $N_c$ unique names in a demographic category $c$, we define the metric $p_{n}^{l} = p(l \vert n)$ for each name $n$. We utilize this metric for our evaluations via various methods described below. 

We first compare the overall accuracy of identifying names as person entity for each demographic category $c$. This is equal to $p_{c}^{l} = \sum_{n \in c} p(n)*p(l \vert n)$.

Next, we compare the distribution of accuracy across all the names of a given demographic. We compare the empirical cumulative density function (ECDF) of the accuracy $p_{n}^{l}$ across all the names $n$ for a given category $c$. This approach allows us to answer the question what percentage of names in a given category have an accuracy lower than $x$. We are particularly interested in observing what percentage of names in a category have an accuracy lower than the accuracy for the OOV name with uninformative embeddings.

In our final comparison, we utilize the confidence estimates of the model (whenever available) for entities which are predicted as person. 
For each name we compute the minimum, mean, median, and standard deviation of the confidence scores. We use these scores to identify the bias in the models.

\section{Results}
\subsection{Overall Accuracy}

\begin{figure*}[!htbp]
    \centering
    \includegraphics[width=\linewidth]{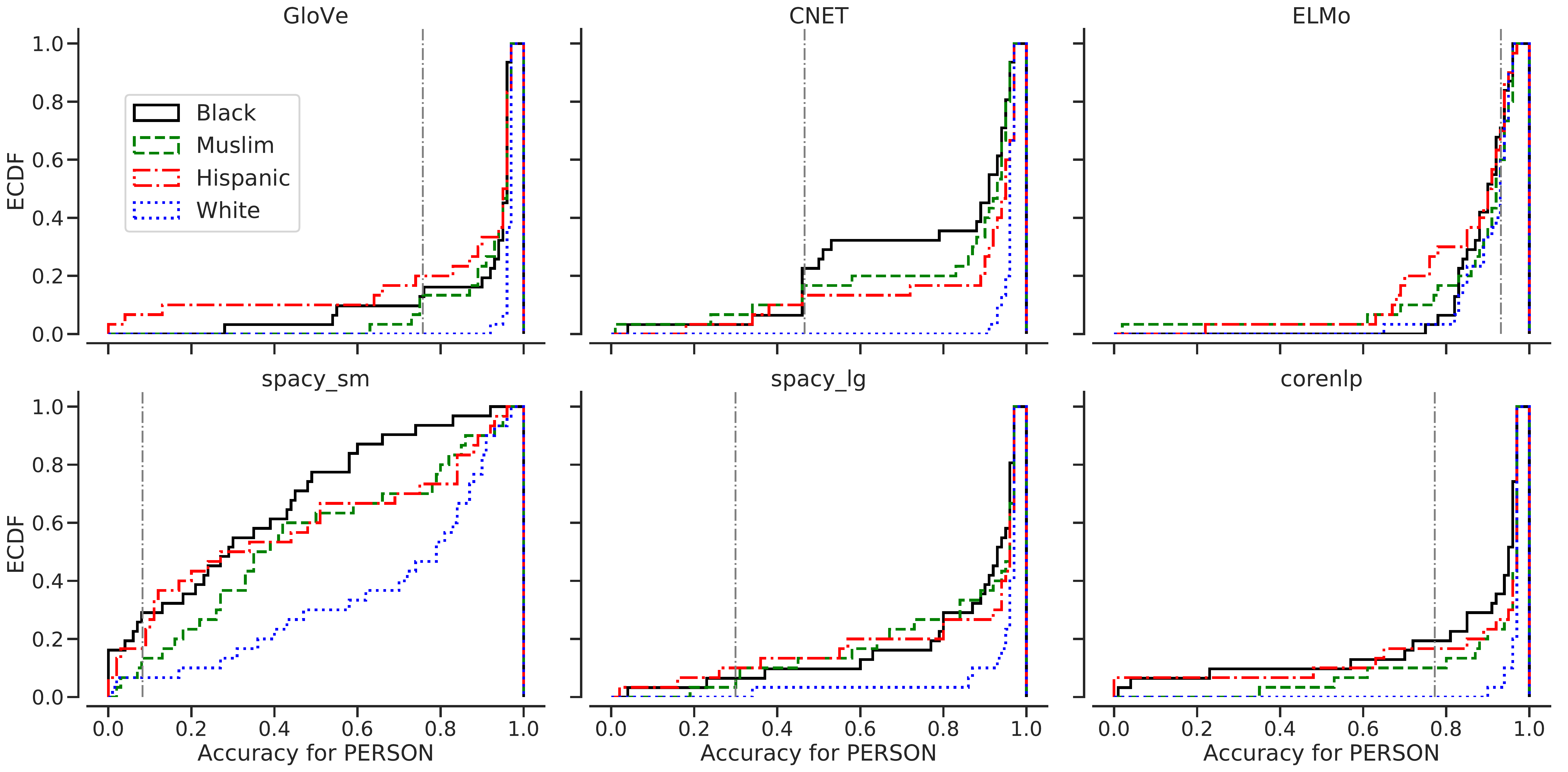}
    \caption{(Best viewed in color) Empirical Cumulative Density Function (ECDF) of names accuracy in Winogender data across demographic categories. The grey vertical line is the confidence percentile for OOV Name. Models with more left skewed accuracy are better (or harder to distinguish plots mean better models).}
    \label{fig:ecdf_all_race}
\end{figure*}

\begin{table*}[!htbp]
    \centering
    \begin{tabular}{lcccccc}
\toprule
\textbf{} & CNET &  ELMo &  GloVe &  corenlp &  spacy\_lg &  spacy\_sm \\
\midrule
\multicolumn{7}{c}{\textbf{WINOGENDER}}\\
\midrule
\textbf{Black Female         } &                      \cellcolor{orange!35} 0.7039 &        0.8942 &                     0.8931 &   \cellcolor{orange!35} 0.7940 &    0.8908 &    \cellcolor{orange!35} 0.3043 \\
\textbf{Black Male         } &                      0.8410 &        0.8986 &                     0.9015 &   0.8862 &    \cellcolor{orange!35} 0.7831 &    0.3517 \\
\midrule
\textbf{Hispanic Female         } &                      0.8454 &        0.8308 &                     0.8738 &   0.8626 &    0.8378 &    0.3726 \\
\textbf{Hispanic Male         } &                      0.8801 &        0.8603 &                     \cellcolor{orange!35} 0.7942 &   0.8629 &    0.8151 &    0.4628 \\
\midrule
\textbf{Muslim Female         } &                      0.8537 &        \cellcolor{orange!35} 0.8130 &                     0.9074 &   0.8747 &    0.8287 &    0.4285 \\
\textbf{Muslim Male         } &                      0.7791 &        \cellcolor{blue!25}0.9265 &                     0.9351 &   0.9477 &    0.8285 &    0.4976 \\
\midrule
\textbf{White Female         } &                      0.9627 &        0.9116 &                     0.9679 &   \cellcolor{blue!25}0.9723 &    \cellcolor{blue!25}0.9577 &    0.5574 \\
\textbf{White Male         } &                      \cellcolor{blue!25}0.9644 &        0.9068 &                     \cellcolor{blue!25}0.9700 &   0.9688 &    0.9260 &    \cellcolor{blue!25}0.7732 \\
\midrule
\textbf{OOV Name       } &                      0.4658 &        0.9318 &                     0.7573 &   0.7724 &    0.2994 &    0.0824 \\
\midrule
\multicolumn{7}{c}{\textbf{IN-SITU}}\\
\midrule

\textbf{Black Female         } &                      \cellcolor{orange!35}0.8289 &        0.8802 &                     0.9193 &   \cellcolor{orange!35}0.8134 &    0.6732 &    \cellcolor{orange!35}0.2104 \\
\textbf{Black Male         } &                      0.8964 &        0.8800 &                     0.9206 &   0.8828 &    0.5922 &    0.2651 \\
\midrule
\textbf{Hispanic Female         } &                      0.8934 &        0.8510 &                     0.9091 &   0.8754 &    0.6736 &    0.3038 \\
\textbf{Hispanic Male         } &                      0.9151 &        0.8729 &                     \cellcolor{orange!35}0.8404 &   0.8699 &    0.6692 &    0.3649 \\
\midrule
\textbf{Muslim Female         } &                      0.9015 &        \cellcolor{orange!35}0.8348 &                     0.9230 &   0.8817 &    \cellcolor{orange!35}0.5686 &    0.3409 \\
\textbf{Muslim Male         } &                      0.8574 &        \cellcolor{blue!25}0.9043 &                     0.9407 &   0.9421 &    0.6890 &    0.4122 \\
\midrule
\textbf{White Female         } &                      \cellcolor{blue!25}0.9619 &        0.8900 &                     \cellcolor{blue!25}0.9555 &   \cellcolor{blue!25}0.9714 &    \cellcolor{blue!25}0.7862 &    0.4503 \\
\textbf{White Male         } &                      0.9541 &        0.8930 &                     0.9504 &   0.9589 &    0.7234 &    \cellcolor{blue!25}0.6388 \\
\midrule
\textbf{OOV Name        } &                      0.7405 &        0.8962 &                     0.8720 &   0.8374 &    0.1003 &    0.0381 \\
\bottomrule
\end{tabular}
    \caption{Overall accuracy for each demographic category, with highlighted \colorbox{blue!25}{best} and \colorbox{orange!35}{worst} performance. We observe significant performance gap between White names and names from other demographics.}
    \label{tab:overall_acc}
\end{table*}

We describe the overall accuracy of various models across demographic categories in Table \ref{tab:overall_acc}. We observe that the accuracy on White names (both male and female) is the highest (except for the ELMo model where the accuracy is highest for Muslim Male names) across all demographic categories and models. We also recognize that the ELMo model exhibits the least variation in accuracy across all demographics, including the OOV names. 
For the ELMo model the bottom three names with the lowest accuracy are Jana (MF), Santiago (HM), and Salma (MF). Among these Jana and Santiago are also most likely to be identified as location entities while Salma is likely to be identified as person entity for $51\%$ cases and location one for $36\%$. 

We observe considerably lower accuracy (3\%-30\%) on uncapitalized names, particularly from the pre-trained CoreNLP and spaCy models such that the bias is no longer evident across the demographic groups (more details in table \ref{tab:overall_acc_lower}). Based on these low accuracy scores, we exclude the results of uncapitalized names in further sections. 
The above results indicate that all considered models are less accurate across non-White names. However, the character embedding based models like ELMo contain the least variation in accuracy across all demographics.

\subsection{Distribution of Accuracy across Names}
Next we look at the distribution of accuracy across names in each demographic category. In Figure \ref{fig:ecdf_all_race}, we report the distribution of name accuracy in Winogender data across all the names in a demographic category for all models. We observe that a large percentage of names from non-White categories have accuracy lower than the OOV names with uninformative embeddings. A similar analysis was conducted for all demographic categories (see figure \ref{fig:ecdf_all}) as well as only for gender categories (see figure \ref{fig:ecdf_all_gender}), but the bias for gender is not as dominant as the other demographic categories. This indicates that the models introduce some biases based on the name's word vector, which causes the lower accuracy of these names. In table \ref{tab:accuracy_range}, we report the variation of accuracy across all names in a given demographic category and confirm that the ELMo model has the least variation. We observe similar results on the In-situ dataset (see figures \ref{fig:ecdf_all_insitu}, \ref{fig:ecdf_gender_insitu}, and \ref{fig:ecdf_race_insitu}).

\begin{table}[]
    \centering
    \begin{tabular}{lrrrr}
    \toprule
    \textbf{model                     } & min$^\star$ & mean$^\star$ &  std$^\dagger$ &   median$^\star$ \\
    \midrule
    \multicolumn{5}{c}{\textbf{WINOGENDER}}\\
    \midrule
    \textbf{CNET} &  0.02 &  0.846 &  0.223 &   0.948 \\
    \textbf{GloVe } &  0.00 &  \textbf{0.903} &  0.170 &   \textbf{0.965} \\
    \textbf{ELMo              } &  \textbf{0.03} &  0.881 &  \textbf{0.126} &   0.922 \\
    \textbf{corenlp                   } &  0.00 &  0.887 &  0.220 &   0.974 \\
    \textbf{spacy\_lg                  } &  0.00 &  0.847 &  0.241 &   0.965 \\
    \textbf{spacy\_sm                  } &  0.00 &  0.460 &  0.327 &   0.425 \\
    \midrule
    \multicolumn{5}{c}{\textbf{IN-SITU}}\\
    \midrule
    \textbf{CNET} &  0.242 &  0.898 &  0.130 &   0.952 \\
    \textbf{GloVe } &  0.159 &  \textbf{0.919} &  0.100 &   0.948 \\
    \textbf{ELMo              } &  \textbf{0.343} &  0.876 &  \textbf{0.067} &   0.889 \\
    \textbf{corenlp                   } &  0.000 &  0.891 &  0.204 &   \textbf{0.969} \\
    \textbf{spacy\_lg                  } &  0.000 &  0.662 &  0.255 &   0.775 \\
    \textbf{spacy\_sm                  } &  0.000 &  0.366 &  0.280 &   0.294 \\
    \bottomrule
    \end{tabular}
    \caption{Range of accuracy values across all names per demographic for each model. Lower is better for $\dagger$  and higher is better for $\star$. }
    \label{tab:accuracy_range}
\end{table}

\subsection{Model Confidence}
\begin{figure*}
    \centering
    \begin{subfigure}{0.325\linewidth}
    \includegraphics[width=\linewidth]{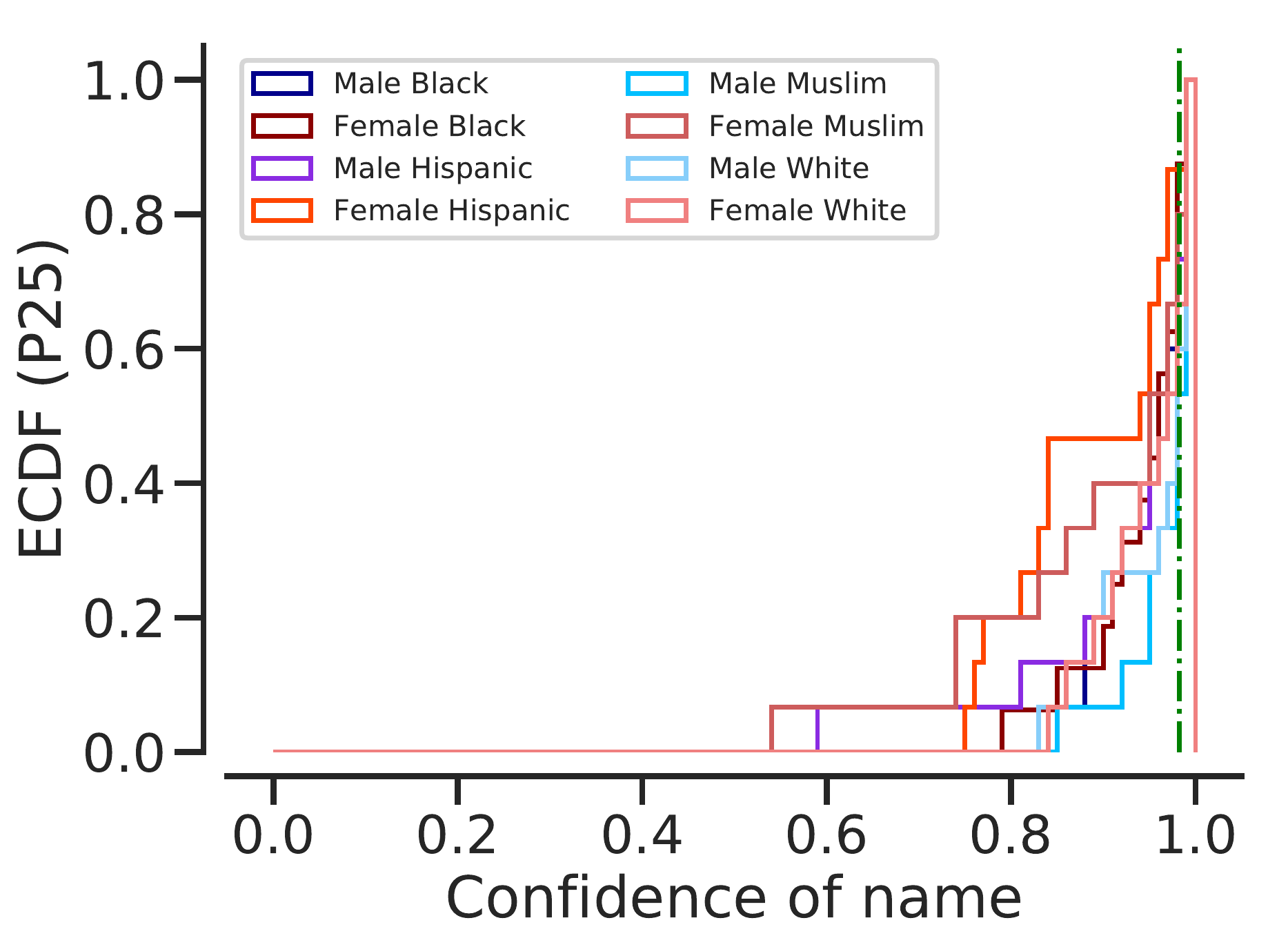}
    \caption{ELMo model}
    \label{fig:confidence_all:elmo}
    \end{subfigure}
    \begin{subfigure}{0.325\linewidth}
    \includegraphics[width=\linewidth]{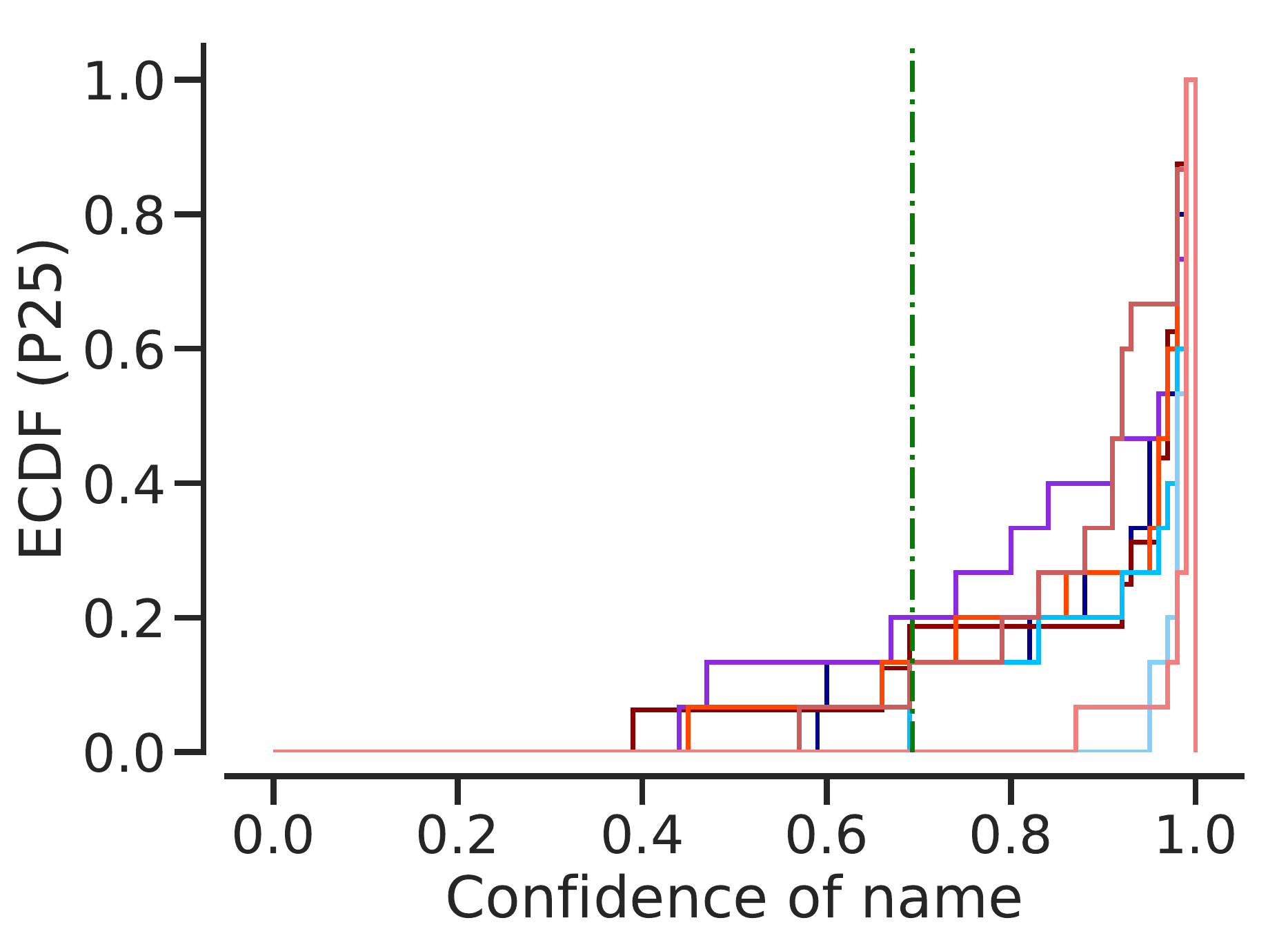}
    \caption{GloVe model}
    \label{fig:confidence_all:glove}
    \end{subfigure}
    \begin{subfigure}{0.325\linewidth}
    \includegraphics[width=\linewidth]{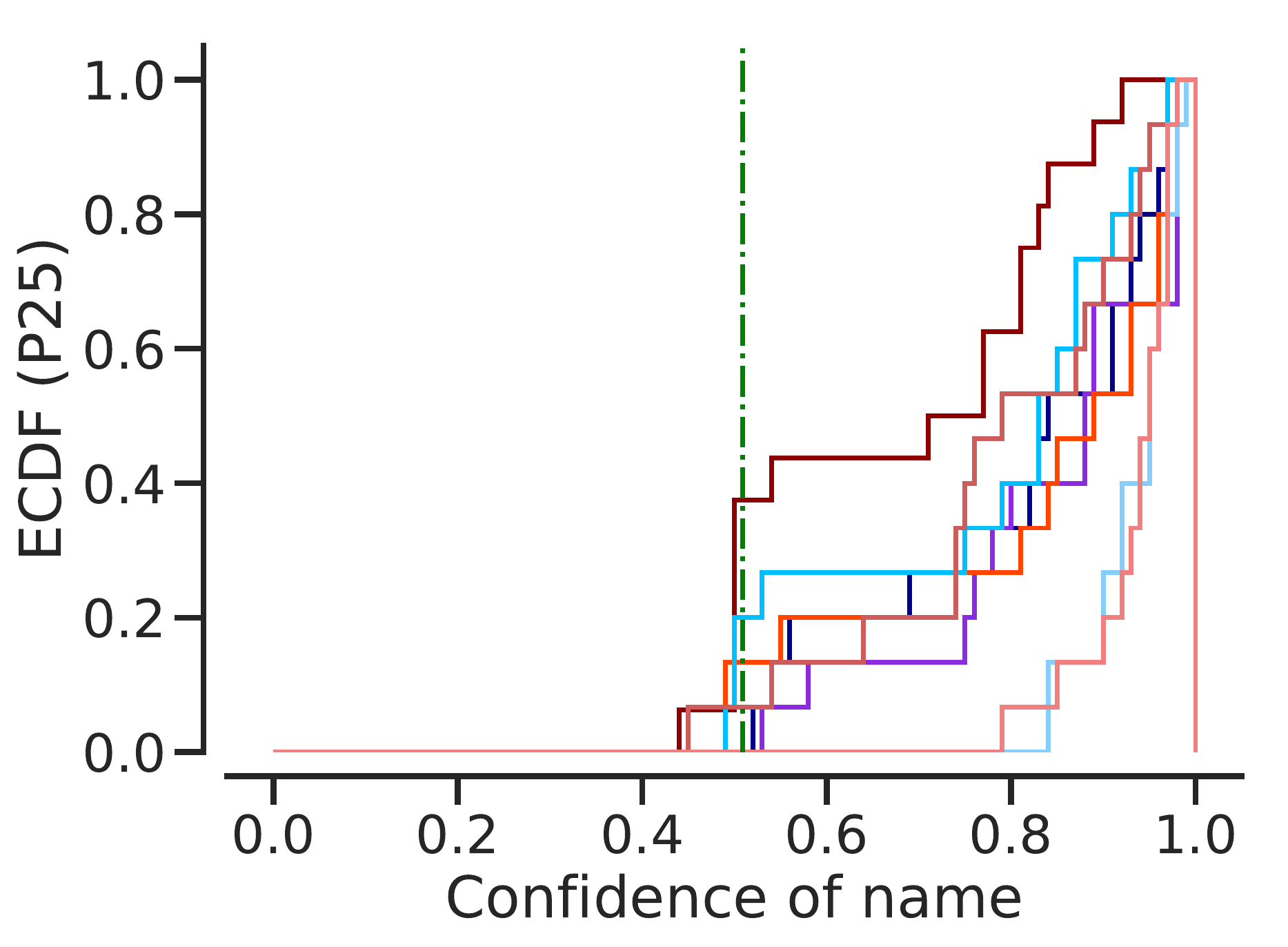}
    \caption{CNET model}
    \label{fig:confidence_all:cnet}
    \end{subfigure}
    \caption{(Best viewed in color) ECDF of percentiles of confidence values for a name to be identified as person entity. The vertical line is the confidence percentile for OOV name baseline. }
    \label{fig:confidence_all}
\end{figure*}

Finally, we investigate the distribution of model confidence across the names which were predicted as person. We use various percentile values for a given name's confidence. We analyze the 25th percentile confidence and the median confidence. As the percentile decreases, the bias observed should become more evident as it highlights the noisier tail of the data. In Figure \ref{fig:confidence_all}, we report the distribution of the 25 percentile values. As before, we observe that a larger percentage of White names have a higher confidence compared to non-White names. Similarly, it can be observed that ELMo based models have the lowest variation in confidence values across all demographics. Surprisingly, the CNET models which are trained on debiased embeddings have the highest variation in confidence estimates. We investigate the variations in median confidence across names in each demographic in Table \ref{tab:confidence_range}. This table confirms our observation above, that ELMo model has least variation across names. We again observe the similar trends for the in-situ data.

\begin{table}[]
    \centering
    \begin{tabular}{lrrrr}
    \toprule
    \textbf{model                     } & min$^\star$ & mean$^\star$ &  std$^\dagger$ &   median$^\star$ \\
    \midrule
    \multicolumn{5}{c}{\textbf{WINOGENDER}}\\
    \midrule
    \textbf{CNET} &  0.495 &  0.894 &  0.132 &   0.956 \\
    \textbf{GloVe } &  0.468 &  0.952 &  0.104 &   0.994 \\
    \textbf{ELMo              } &  \textbf{0.621} &  \textbf{0.980} &  \textbf{0.046} &   \textbf{0.995} \\
    \midrule
    \multicolumn{5}{c}{\textbf{IN-SITU}}\\
    \midrule
    \textbf{CNET} &  0.606 &  0.946 &  0.080 &   0.981 \\
    \textbf{GloVe } &  0.668 &  0.983 &  0.049 &   \textbf{0.998} \\
    \textbf{ELMo              } &  \textbf{0.831} &  \textbf{0.994} &  \textbf{0.017} &   \textbf{0.998} \\
    \bottomrule
    \end{tabular}
    \caption{Range of median confidence values across all names per demographic for each model. Confidence values unavailable for other models. Lower is better for $\dagger$  and higher is better for $\star$ }
    \label{tab:confidence_range}
\end{table}

\section{Discussion}
\label{discussion}

\begin{figure*}[!htbp]
    \centering
    \includegraphics[width=\linewidth]{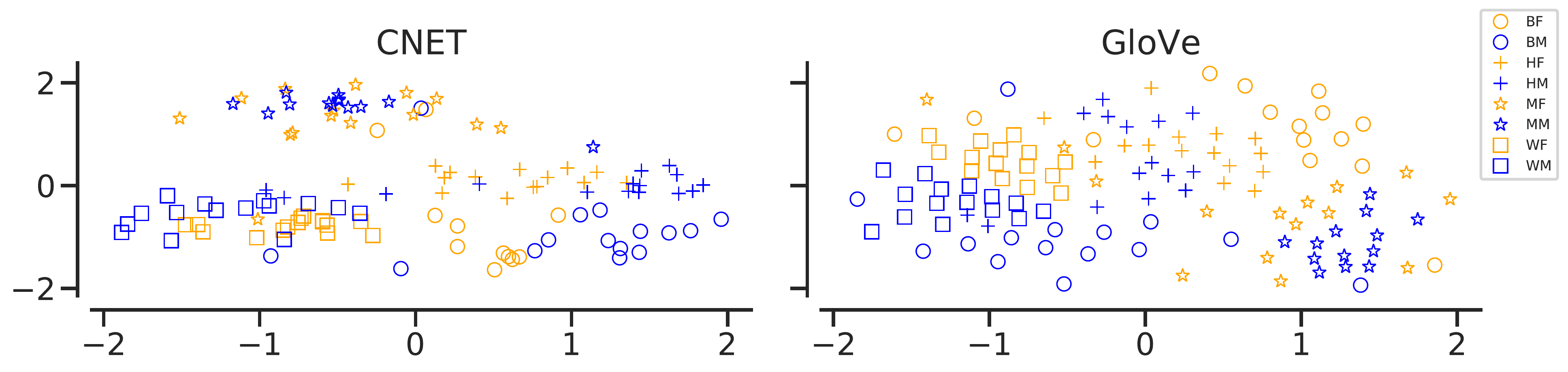}
    \caption{t-SNE projections of first name embeddings identified by their demographic categories (best viewed in color).}
    \label{fig:name_embedding}
\end{figure*}

Our work sheds light on the variation in accuracy of named entity recognition systems on first names which are prominent, in certain demographic categories such as gender and race. A lower dimension projection (obtained via t-SNE) of the embeddings as shown in Figure \ref{fig:name_embedding}) reveals that the name embeddings do cluster based on their demographic information. The clustering is more prominent across the race dimension. 

It is important to note that the performance gap between names from different demographic groups can be partially attributed to the bias in the training data. Built from the Reuters 1996 news corpus, CoNLL03 is one of the most widely-used NER dataset. However, as shown in Table \ref{tab:name_category_frequency}, the CoNLL03 training data contains significantly more Male names than Female names and more White names than non-White names.  

While this work has approached studying the issue of bias using a synthetic dataset, it is still helpful in uncovering various aspects of the NER pipeline. We specifically identified variation in NER accuracy by using different embeddings. This is important because NER facilitates multiple automated systems, e.g. knowledge base construction, question answering systems, search result ranking, and automated keyword identification. If named entities from certain parts of the populations are systematically misidentified or mislabeled, the damage will be twofold: they will not be able to benefit from online exposure as much as they would have if they belonged to a different category (Allocation Bias \footnote{\url{https://catalogofbias.org/biases/allocation-bias/}} as defined in  \cite{Barocas2017Crawford}) and they will be less likely to be included in future iterations of training data therefore perpetuating the vicious cycle (Representation bias). Furthermore, while a lot of research in bias has focused on just one  aspect of demographics (i.e. only race or only gender) our work focuses on the intersectionality of both these factors. Similar research in the domain of bias across gender, ethnicity, and nationality has been studied in bibliometric literature \cite{Mishra_2018}.

\textbf{Limitations} Our current work is limited in its analysis to only unigram entities. A major challenge for correctly constructing and evaluating our methods for n-gram entities is to come up with a collection of names which are representative of demographics. While first name data is easily available through various census portals, full name data tagged with demographic information is harder to find. Furthermore, when extending this analysis to n-gram entities we need to define better evaluation metrics, i.e. how different is a mistake on the first name from a mistake on other parts of the name, and how to quantify this bias appropriately. Finally, we are aware that our name lists are based on old data and certain first names are be more likely to be adopted by other communities, leading to the demographic association of names to change across time \cite{Smith2013Genni}. However, these factors do not affect our analysis as our name collection consists of dominant names in a demographic. Additionally, our work can be extended to other named entity categories like location, and organizations from different countries so as to assess the bias in identifying these entities. Since, our analysis focused on NER models trained on English corpus, another line of research will be to see if models trained in other languages also contain favorable results for named entities more likely to be used in cultures where that language is popular. This should lead to the assessment of NER models in different languages with named entities representing a larger demographic diversity. Finally, the goal of this paper has been to identify biases in accuracy of NER models. We are investigating ways to mitigate these biases in an efficient manner.

\section{Related work}

Bias in embeddings has been studied by \citet{Bolukbasi2016}, who showed that the vector for stereotypically male professions are closer to the vector for “man” than “woman” (e.g. “Man is to Computer Programmer as Woman is to Homemaker”). Techniques to debias embeddings were suggested, where a “gender” direction is identified in the vector space and thus subtracted from the embeddings.  More recently  \citet{gonen-goldberg-2019-lipstick-pig}  showed how those efforts are not substantially removing bias, rather hiding it: words with similar biases are still clustered together in the de-biased space.
\citet{Manzini2019BlackIT} extended the techinques of \cite{Bolukbasi2016} to multi-class setting, instead of just binary ones. Emebddings were also the subject of scrutiny in \citet{Caliskan2017SemanticsDA}, where a modified version of the implicit association tests \cite{Greenwald1998MeasuringID} were developed.

The Winogender schemas we used in this works were developed by \citep{rudinger-etal-2017-social} to study gender bias in coreference resolution.

\section{Conclusion}
In this work, we introduced a novel framework to study the bias in named entity recognition models using synthetically generated data. From our analysis reports that models are better at identifying White names across all datasets with higher confidence compared with other demographics such as Black names. We also demonstrate that debiased embeddings do not help in resolving the bias in recognizing names. Finally, our results show that character based models, such as ELMo, result in the least bias across demographic categories, but those models are still unable to entirely remove the bias. Since, NER models are often the first step in automatic construction of knowledge bases, our results can help identify potential issues of bias in KB constructions.

\begin{acks}
\end{acks}

\bibliographystyle{ACM-Reference-Format}
\bibliography{sample-base}

\appendix
\onecolumn

\section{Appendix}
\subsection{Name distribution in data}

\begin{table*}[!htbp]
    \centering
    %\begin{tabular}{p{0.37\linewidth}p{0.2\linewidth}p{0.36\linewidth}}
    \begin{tabular}{lrl}
\toprule
\textbf{Category} & \textbf{Total Count} & \textbf{Most Common Name (Count)}\\
\midrule
\textbf{Black Female (BF)      } &  0 & --  \\
\textbf{Black Male (BM)      } &  18 & Malik (13) \\
\midrule
\textbf{Hispanic Female (HF) } & 22 & Maria (12) \\
\textbf{Hispanic Male (HM)      } &  89 & Jose (20) \\
\midrule
\textbf{Muslim Female (MF)      } & 8 & Jana (6) \\
\textbf{Muslim Male (MM)      } & 68 & Ahmed (49) \\
\midrule
\textbf{White Female (WF)      } &  17 & Stephanie (6) \\
\textbf{White Male (WM)      } &  148 & Paul (51) \\
\bottomrule
\end{tabular}
    \caption{Name distribution in CoNLL03 training data across different categories}
    \label{tab:name_category_frequency}
\end{table*}

\subsection{Distribution of accuracy for various subsets of data for Winogender analysis}

\begin{table*}[!htbp]
    \centering
    \begin{tabular}{lcccccc}
\toprule
\textbf{} & CNET &  ELMo &  GloVe &  corenlp &  spacy\_lg &  spacy\_sm \\
\midrule
\multicolumn{7}{c}{\textbf{WINOGENDER LOWER}}\\
\midrule

\textbf{Black Female         } &                            \cellcolor{orange!35}0.0018 &              0.8695 &                           \cellcolor{orange!35}0.6855 &         0.0230 &          \cellcolor{orange!35}0.0915 &             NaN \\
\textbf{Black Male         } &                            \cellcolor{blue!25}0.0911 &              0.8764 &                           0.8068 &         0.0292 &          0.2077 &             NaN \\
\midrule
\textbf{Hispanic Female         } &                            0.0572 &              0.8137 &                           0.7624 &         \cellcolor{blue!25}0.0581 &          0.1496 &             NaN \\
\textbf{Hispanic Male         } &                            0.0556 &              0.8401 &                           0.7408 &         0.0321 &          \cellcolor{blue!25}0.3044 &             NaN \\
\midrule
\textbf{Muslim Female         } &                            0.0192 &              \cellcolor{orange!35}0.7982 &                           0.7517 &         0.0164 &          0.1797 &             NaN \\
\textbf{Muslim Male         } &                            0.0222 &              \cellcolor{blue!25}0.9031 &                           0.8118 &         \cellcolor{orange!35}0.0088 &          0.2787 &             NaN \\
\midrule
\textbf{White Female         } &                            0.0288 &              0.8779 &                           \cellcolor{blue!25}0.8363 &         0.0552 &          0.1385 &      0.0000 \\
\textbf{White Male         } &                            0.0318 &              0.8736 &                           0.7839 &         0.0193 &          0.2920 &             NaN \\
\midrule
\textbf{OOV Name        } &                               NaN &              0.9256 &                           0.0001 &            NaN &             NaN &             NaN \\
\midrule
\multicolumn{7}{c}{\textbf{IN-SITU LOWER}}\\
\midrule
\textbf{Black Female         } &                            \cellcolor{orange!35}0.0087 &              0.8774 &                           \cellcolor{orange!35}0.7855 &         0.0151 &          0.0519 &             NaN \\
\textbf{Black Male         } &                            0.1679 &              0.8759 &                           0.8895 &         0.0291 &          0.0877 &             NaN \\
\midrule
\textbf{Hispanic Female         } &                            0.1066 &              0.8482 &                           0.8750 &         \cellcolor{blue!25}0.0678 &          0.0634 &             NaN \\
\textbf{Hispanic Male         } &                            \cellcolor{blue!25}0.1137 &              0.8697 &                           0.8226 &         0.0429 &          \cellcolor{blue!25}0.1712 &             NaN \\
\midrule
\textbf{Muslim Female         } &                            0.0480 &              \cellcolor{orange!35}0.8332 &                           0.8706 &         0.0136 &          0.1045 &             NaN \\
\textbf{Muslim Male         } &                            0.0544 &              \cellcolor{blue!25}0.8987 &                           0.8517 &         \cellcolor{orange!35}0.0065 &          0.1453 &             NaN \\
\midrule
\textbf{White Female         } &                            0.0826 &              0.8844 &                           \cellcolor{blue!25}0.9340 &         0.0544 &          \cellcolor{orange!35}0.0388 &          0.0005 \\
\textbf{White Male         } &                            0.0867 &              0.8872 &                           0.9059 &         0.0418 &          0.1398 &             NaN \\
\midrule
\textbf{OOV Name        } &                               NaN &              0.8962 &                           0.2353 &            NaN &             NaN &             NaN \\

\bottomrule
\end{tabular}
    \caption{Overall accuracy on lower cased data for each demographic category, with highlighted \colorbox{blue!25}{best} and \colorbox{orange!35}{worst} performance.}
    \label{tab:overall_acc_lower}
\end{table*}

\begin{figure*}[!htbp]
    \centering
    \includegraphics[width=\linewidth]{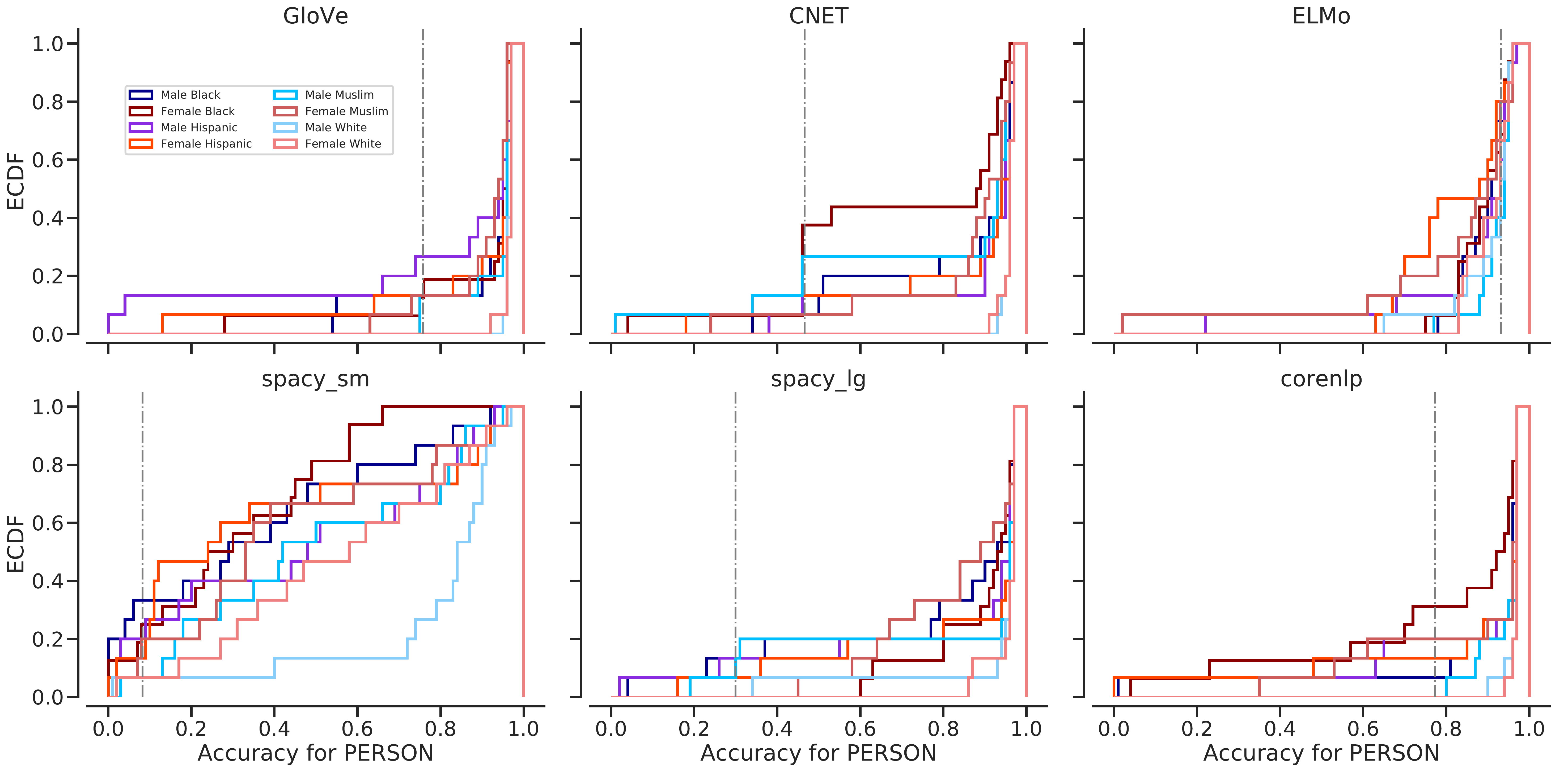}
    \caption{(Best viewed in color) Empirical Cumulative Density Function (ECDF) of names accuracy in Winogender data across demographic categories. The grey vertical line is the confidence percentile for OOV Name. Models with more left skewed accuracy are better (or harder to distinguish plots mean better models).}
    \label{fig:ecdf_all}
\end{figure*}

\begin{figure*}[!htbp]
    \centering
    \includegraphics[width=\linewidth]{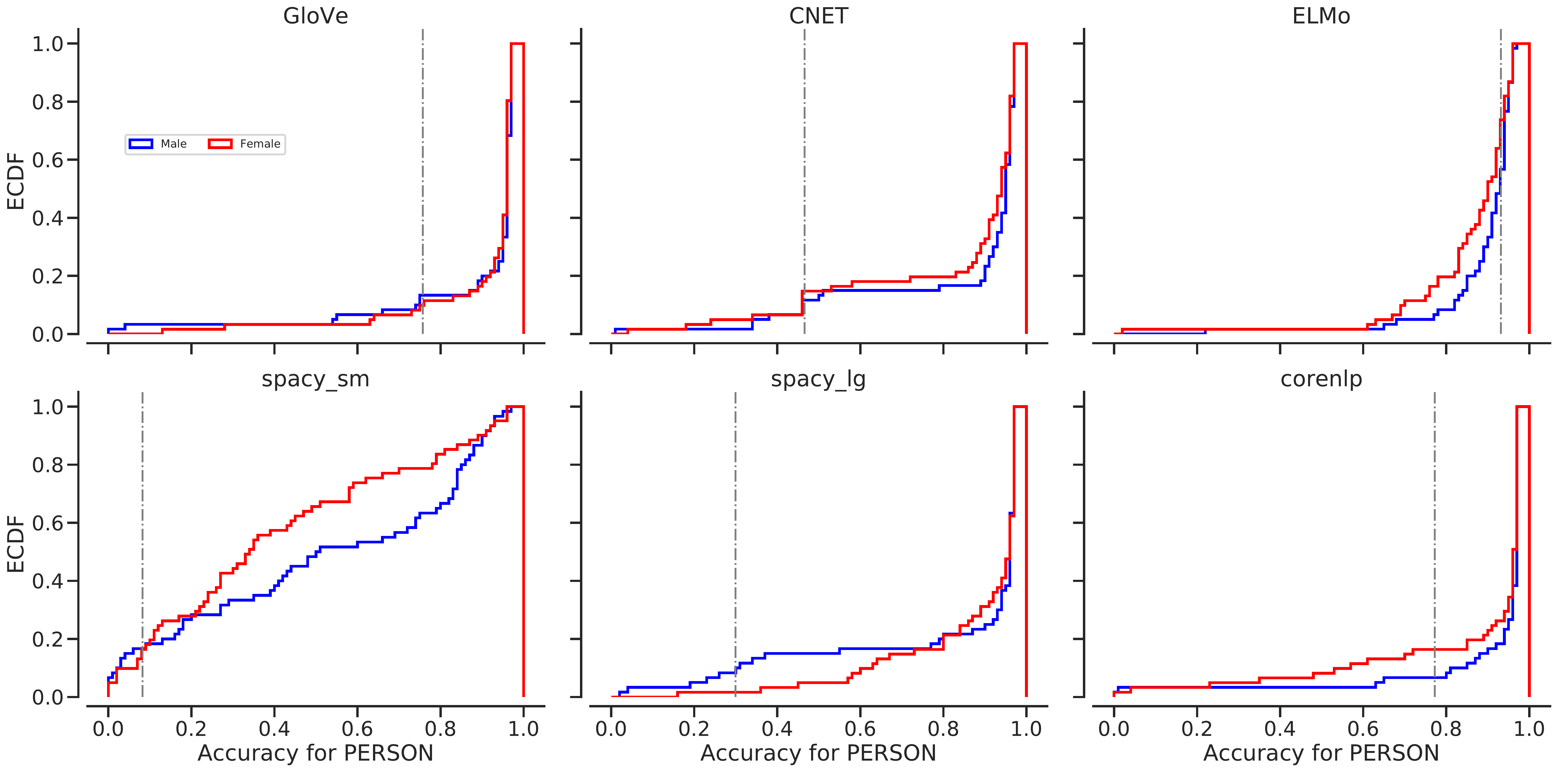}
    \caption{(Best viewed in color) Empirical Cumulative Density Function (ECDF) of names accuracy in Winogender data across demographic categories. The grey vertical line is the confidence percentile for OOV Name. Models with more left skewed accuracy are better (or harder to distinguish plots mean better models).}
    \label{fig:ecdf_all_gender}
\end{figure*}

% \subsection{Distribution of accuracy for various subsets of data for In-Situ analysis}

\begin{figure*}[!htbp]
    \centering
    \includegraphics[width=\linewidth]{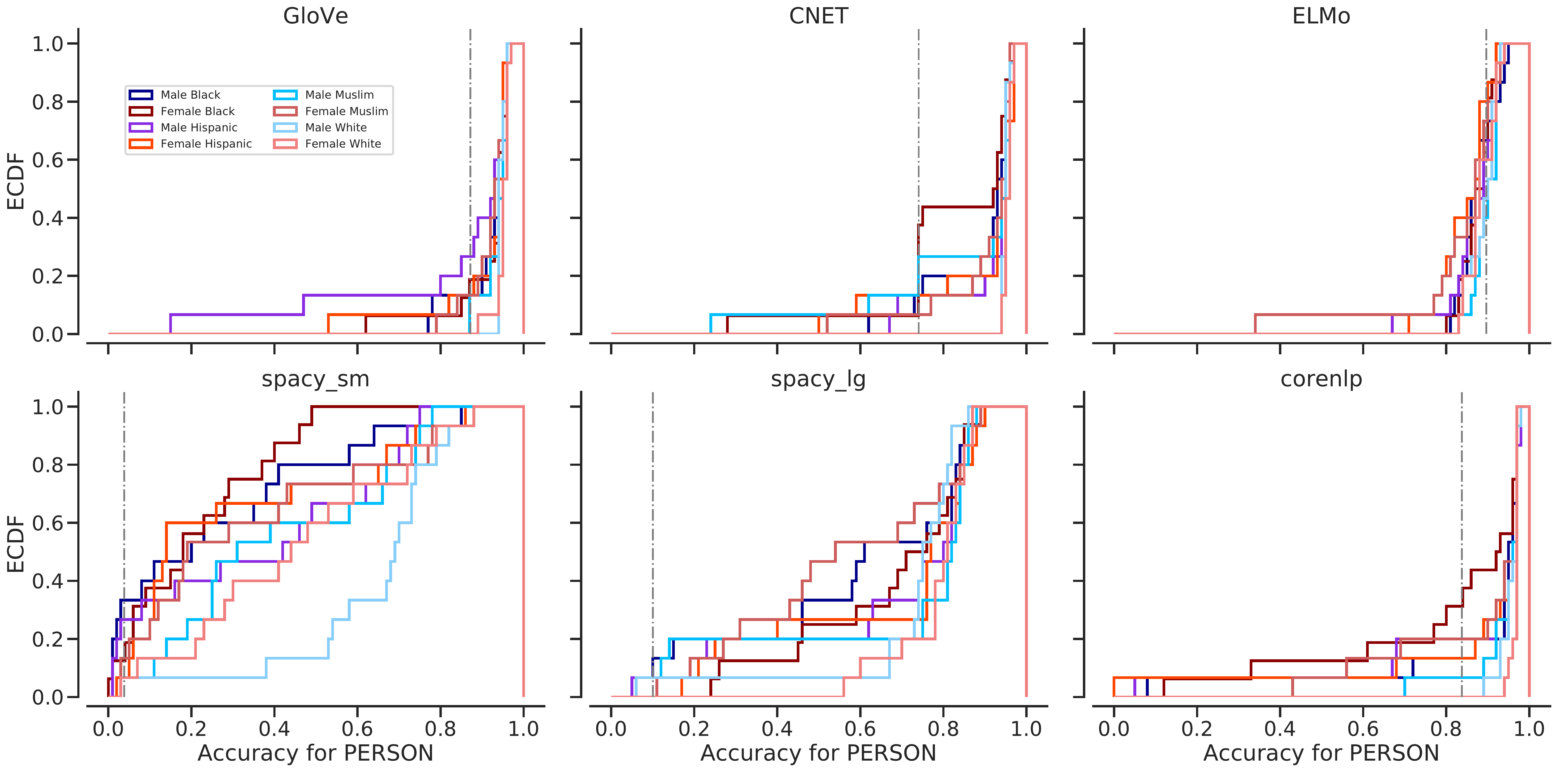}
    \caption{(Best viewed in color) Empirical Cumulative Density Function (ECDF) of names accuracy in In-Situ data across demographic categories. The grey vertical line is the confidence percentile for OOV Name. Models with more left skewed accuracy are better (or harder to distinguish plots mean better models).}
    \label{fig:ecdf_all_insitu}
\end{figure*}

\begin{figure*}[!htbp]
    \centering
    \includegraphics[width=\linewidth]{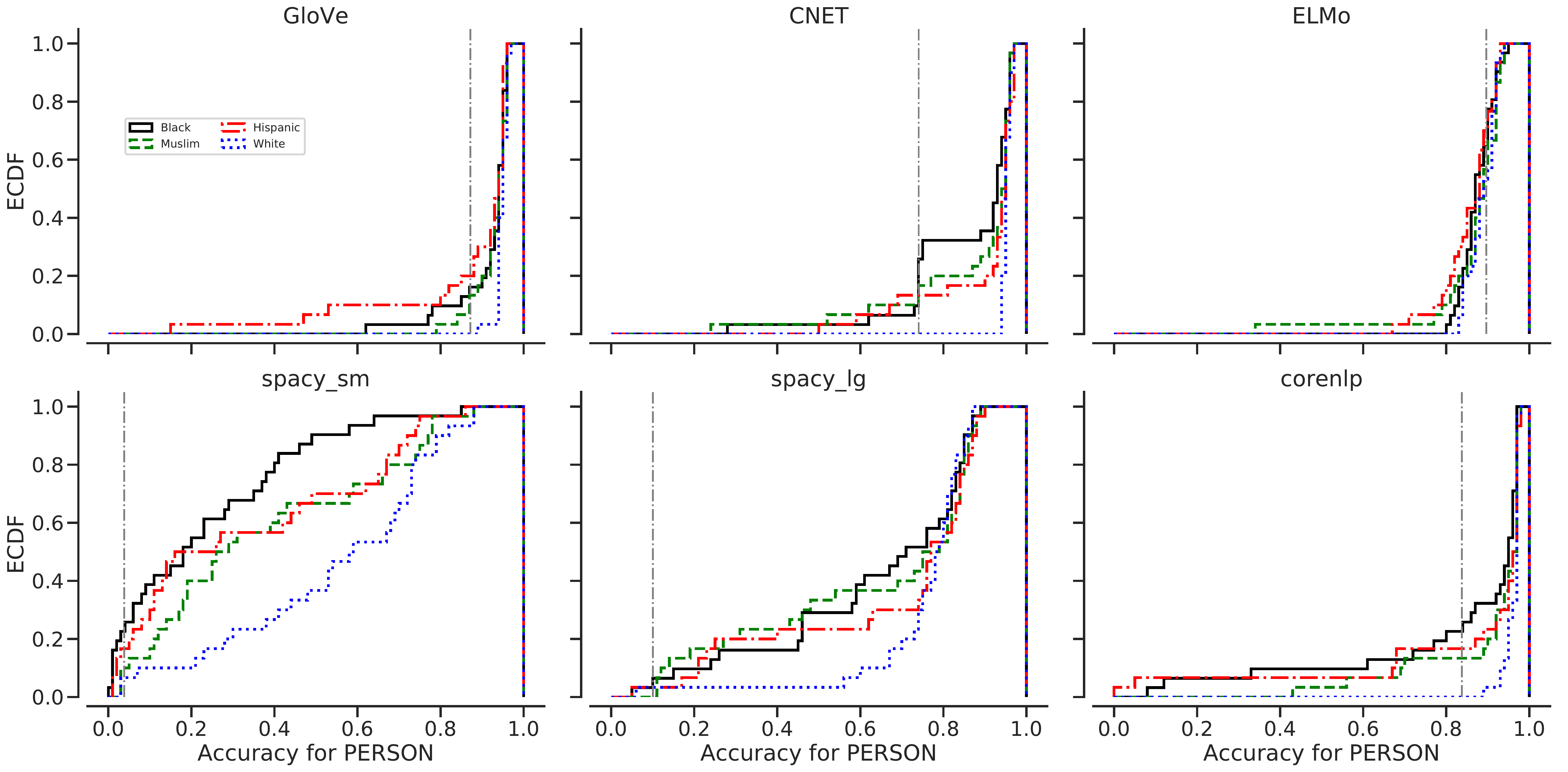}
    \caption{(Best viewed in color) Empirical Cumulative Density Function (ECDF) of names accuracy in In-Situ data across demographic categories. The grey vertical line is the confidence percentile for OOV Name. Models with more left skewed accuracy are better (or harder to distinguish plots mean better models).}
    \label{fig:ecdf_race_insitu}
\end{figure*}

\begin{figure*}[!htbp]
    \centering
    \includegraphics[width=\linewidth]{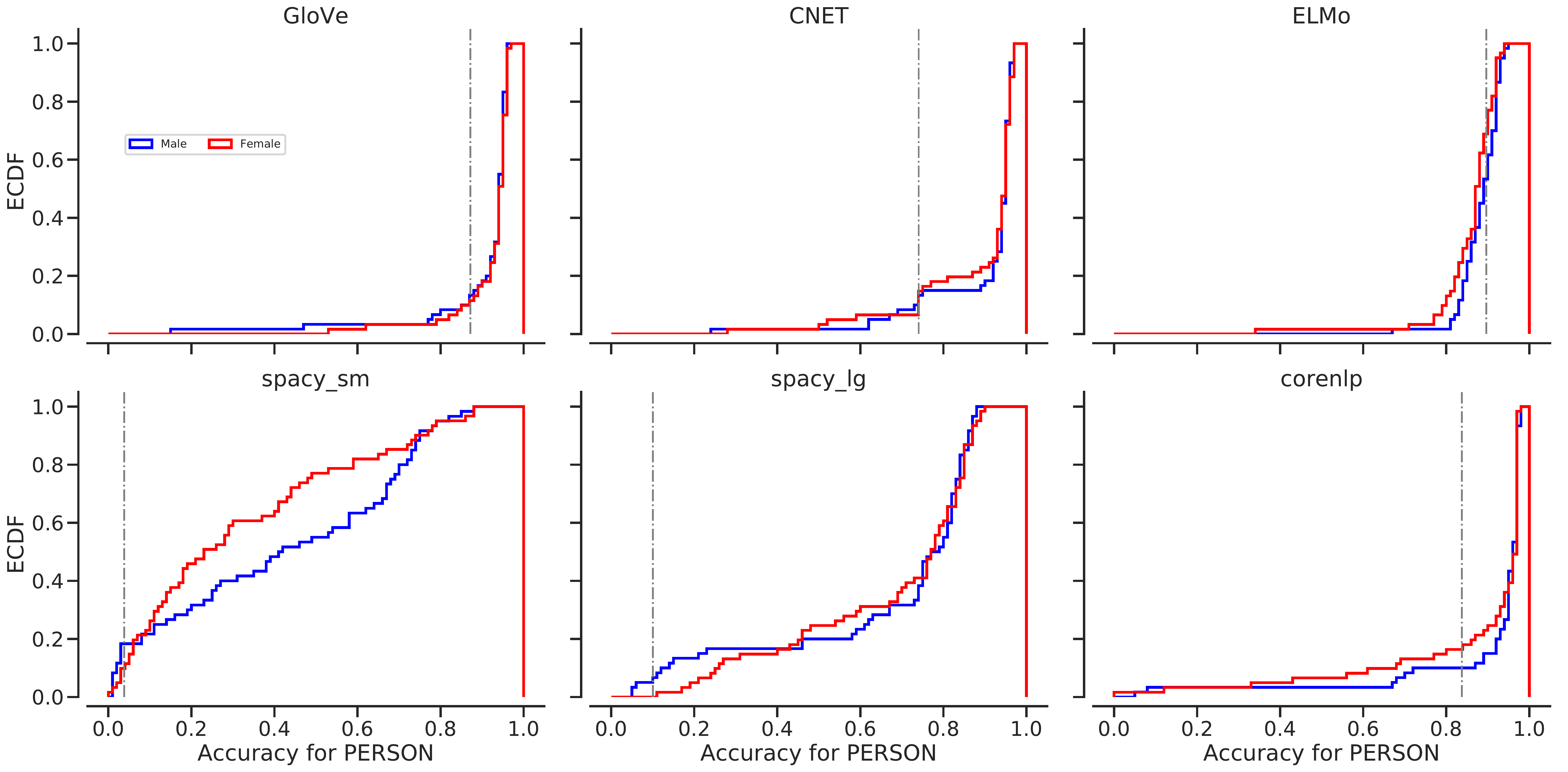}
    \caption{(Best viewed in color) Empirical Cumulative Density Function (ECDF) of names accuracy in In-Situ data across demographic categories. The grey vertical line is the confidence percentile for OOV Name. Models with more left skewed accuracy are better (or harder to distinguish plots mean better models).}
    \label{fig:ecdf_gender_insitu}
\end{figure*}








\end{document}